# Generating synthetic multi-dimensional molecular-mediator time series data for artificial intelligence-based disease trajectory forecasting and drug development digital twins: Considerations

Gary An[1*] and Chase Cockrell[1]
1. University of Vermont Larner College of Medicine

Keywords: Synthetic data, molecular time series data, agent-based model, artificial intelligence, machine learning, artificial neural network, multiscale model, medical digital twin

*Corresponding Author:
89 Beaumont Ave
Given 319D
Burlington, VT 05405
gan@med.uvm.edu

## Abstract

The use of synthetic data is recognized as a crucial step in the development of neural network-based Artificial Intelligence (AI) systems. While the methods for generating synthetic data for AI applications in other domains have a role in certain biomedical AI systems, primarily related to image processing, there is a critical gap in the generation of time series data for AI tasks where it is necessary to know how the system works. This is most pronounced in the ability to generate synthetic multi-dimensional molecular time series data (SMMTSD); this is the type of data that underpins research into biomarkers and mediator signatures for forecasting various diseases and is an essential component of the drug development pipeline. We argue the insufficiency of statistical and data-centric machine learning (ML) means of generating this type of synthetic data is due to a combination of factors: perpetual data sparsity due to the Curse of Dimensionality, the inapplicability of the Central Limit Theorem in terms of making assumptions about the statistical distributions of this type of data, and the Causal Hierarchy Theorem, which intrinsically limits the ability of data-centric methods to make statements about generative mechanisms that cross-scales (as is the case from cellular-molecular biology to an individual person's state of health and disease). Alternatively, we present a rationale for using complex multi-scale mechanism-based simulation models, constructed and operated on to account for perpetual epistemic incompleteness and the need to provide maximal expansiveness in concordance with the Principle of Maximal Entropy. These procedures provide for the generation of SMMTD that minimizes the known shortcomings associated with neural network AI systems, namely overfitting and lack of generalizability. The generation of synthetic data that accounts for the identified factors of multi-dimensional time series data is an essential capability for the development of mediator-biomarker based AI forecasting systems, and therapeutic control development and optimization through systems like Drug Development Digital Twins.

## 1.0 Introduction: What this article is, and is not, about

Synthetic data is recognized to be needed for machine learning (ML) and artificial intelligence (AI) systems to reach their full potential [1]. While many of these concepts can find application in biomedical AI, this article is specifically about generating synthetic multi-dimensional molecular time series data (SMMTSD) for biomedical applications where it is critical *to have an understanding of how the system works*. These include the use of molecular-level biomarker/-omics panels to forecast the behavior of an individual patient or the evaluation of whether novel drugs or novel combinations of existing drugs (repurposing) will work; note that this is distinct from the task of searching for a particular candidate compound based on a presumed target pathway/gene/molecule. These forecasting or evaluative tasks presume a generative causal relationship between the lower scale features (cellular/molecular) and the higher order, system-level phenotype; these are multi-scale, hierarchical causal tasks [2] distinct from producing and using structural causal models whose feature sets are scale-agnostic and flat [3]. The focus on cellular-molecular biology places these systems apart from those that can be readily characterized by physics-based equations, reflecting the vast majority of biomedical research on the discovery and development of potential drugs. We pose that generating this type of multi-scale, hierarchical time series data precludes the use of: 1) statistical methods, including the use of generative adversarial neural networks (GANs) and 2) existing approaches that generate virtual/synthetic populations that, while producing heterogeneous synthetic populations, do so by aggregating classes of patients around mean mediator values and do not replicate individual patient trajectories [4-7]. We further propose that these tasks require the use of sufficiently complex, mechanism-based simulation models of cellular and molecular processes to generate SMMTSD to train artificial neural networks (ANNs)/artificial intelligence (AI) systems to address these tasks.

## 2.0 Synthetic Data: Current uses and means of generation

Training ANNs for AI applications is notoriously data hungry, and even in areas with copious data there are recognized benefits to the supplementation of that data with synthetically generated data [1]. The most well-publicized examples of AI use synthetic data: image recognition/generation [8], natural language processing/generation [9], self-driving cars [10] and game-playing systems [11-13]. In fact, the development of AI image generators and chatbots integrates the generation of synthetic data with the target applications, which are essentially means of generating "realistic" synthetic objects. The means of generating synthetic data into two general groups:

1. *Statistical* synthetic data can be generated if there is enough existing real-world data such that either: 1) the statistical distribution of system features can be reliably approximated or 2) an ANN can be trained, based on their property as Universal Approximators [14], to a sufficiently robust generative function through the iterative use of training various combinations, often including generative adversarial neural networks or GANNs [15, 16]. However, if an ANN is going to be used to approximate the generative function underlying the data, there needs to be sufficient existing training data such that the iterative process provided by an adversarial network training scheme can get started. This property is evident in the two most successful applications of this type, image recognition/generation (for example, Dall-E2 https://openai.com/product/dall-e-2 and Midjourney https://www.midjourney.com/home/) and language processing/generation (for example, Chat-GPT https://chat-gpt.org/).

2. *Simulation* generated data from mechanism-based simulation models can be considered "real enough" if the generative simulations are firmly grounded in natural laws ("physics-based") or within the context of known rules (e.g. games). Here there is a high degree of confidence in the rules and mechanisms of the simulations and thus concurrent trust in the fidelity of the synthetically generated data and the "real-world" in which the trained systems must operate (acknowledging that in the case of a game, the game itself represent the "real world" for the player).

## 3.0 Biomedical Synthetic Data: Cases where existing methods can, and cannot, be used

There are biomedical use cases where the above well-vetted means of generating synthetic data through statistical or physics-based methods can be applied [17]. Biomedical image processing (for either radiology or pathology) readily falls into the category of general image processing, and the same approaches used for image analysis can be extrapolated to biomedical applications [17-23]. Other circumstances where there may be enough existing data for statistical distributions are population-level data suitable to represent the control population in a potential clinical trial [24-26] or based on data from electronic health records [27-32]. In terms of simulated synthetic data, biomedical systems that can be represented as physical systems, such as fluid dynamics for anatomic representation of blood flow , electrical circuits for cardiac conduction, or the mechanical properties of joints, can be simulated with well-recognized "physics-based" methods [33-36].

However, for the vast majority of biomedical research, namely experimental cellular and molecular biology, neither of these conditions hold. A central premise of experimental biomedical research is that more granular mechanistic knowledge can lead to improved human health, i.e. through the development and use of various -omics-based and multiplexed molecular assays. The entire drug development endeavor is based on the premise that more detailed molecular knowledge of biological processes is the means to identifying more effective and precise new therapeutic agents. However, this experimental paradigm has two consequences that challenge the application of statistical, data-centric forms of analysis and, consequently, the ability to generate synthetic data.

1. Perpetual Data Sparsity. As each new "feature" (gene, biomarker, etc.) is identified, this adds dimensionality to the characterization of biological systems and this process carries with it a cost: the Curse of Dimensionality [37]. The Curse of Dimensionality means that with each additional feature used to describe a system, there is an exponential increase in the potential configurations those features can take relative to each other (combinatorial explosion) and, similarly, the amount of data/sample points needed to characterize those potential combinations; *this leads to a state of perpetual data sparsity*. Because the space of potential combinations is perpetually under-sampled in the real world, it is not possible to generate the "true" statistical distribution of these values. Also, since the variables are not independent (e.g. by virtual of the existence of complex biological networks), the Central Limit Theorem, which allows for the assumption of normal distributions, does not apply. The fact that the variance of these types of measurement is not normally distributed is immediately evident when one examines multi-molecular time series data. Furthermore, it is readily evident that the shape of the probability distribution can vary from time point to time point;

this further limits the potential application of any constant noise term to the probability distribution of time series values (as would be commonly used in stochastic differential equations). Therefore, irrespective of the fundamental source of data heterogeneity and variation, be it from epistemic stochasticity or aleatory (intrinsic) stochasticity, the noise function for multiplexed molecular time series data is a complex function that cannot be extracted from the data alone.

2. Hierarchical Causality. The drug discovery endeavor premised on finding potential interventions (drugs) at the molecular level that can then affect total system phenotype (e.g. disease state) is inherently hierarchical and multiscale. The hierarchical nature of this task means that it is subject to the limits of the Causal Hierarchy Theorem, which states that no data-driven approach can identify generative causal models that cross multiple scales [2, 38]. Hierarchical causality is distinct from structural causal models that treat causal features in a scale agnostic fashion, as is the case in many ML classification, optimization or forecasting tasks; for example, a ML derived predictive algorithm that utilizes a scale-agnostic feature set including demographic and clinical data, physiological signals, laboratory tests or non-time series molecular/-omics profiles [25, 26, 39]. In these cases the resulting structural causal model would operate on statistical/correlative causality in terms of likelihood of events following each other [3], but there is not presumed generative, mechanistic link in the time course of these events (and is therefore not hierarchical (15)). The need to account for generative hierarchical causality when evaluating the relationship between cellular-molecular processes and how they manifest at the individual patient (system level) severely limits (prevents) the use of formal causal inference for this task. We note that the challenge of demonstrating generative hierarchical causality is the primary bottleneck in the translation of the output of cellular/molecular biology into therapeutics that can modulate those cellular and molecular behaviors in a way that demonstrates clinical effectiveness: the Translational Dilemma [40, 41].

Because statistical/data-centric approaches have the above fundamental limitations in terms of generating SMMTSD, mechanism-based simulation methods need to be applied. However, there are also specific issues to directly translating the concept of "physics-based" simulations to those that can be used to represent the multiscale effects of cellular/molecular biology.

1. Multiscale cellular/molecular biology biology simulations cannot be produced *ab initio*. While biology, as a physical system, is certainly bound by the fundamental laws of physics and chemistry, there are no corresponding fundamental laws that constrain the dynamics and output of cellular/molecular biology. While certain physical laws and constraints can be incorporated into cellular/molecular simulations (i.e., mechanical effects on cell signaling or mass conservation in metabolism), these components must inevitably be connected to representations of how cells respond to and modulate their behavior to these features; it is these behavioral aspects of cellular/molecular biology that produce the richness of biology, and for which there are not constraining fundamental laws. Because of this, the essential features of cellular/molecular biology cannot be represented by "physics-based" simulations [33-36]. Ultimately, the fundamental problem is that the generative mechanisms that lead to molecular mediator time series data are mostly unknown; further, they are undiscoverable (at present) by theory because the set of events that generate molecular mediator time series data are not tractable to compute *ab initio* (i.e., from the electronic configurations of the relevant molecules).

2. Dealing with the impact of perpetual epistemic uncertainty and incompleteness. There is perpetual epistemic uncertainty in terms of the molecular-cellular rules that govern the biological system. Pushing the boundary of knowledge is the entire goal of cellular/molecular biology, but it must be acknowledged that it is impossible to know everything. Therefore, there must be some way of using incomplete knowledge to in a useful fashion. The only means of accomplishing this is by posing a particular mechanistic hypothesis and then operating on that hypothesis structure (through iterative experiment, recalibration/validation and iterative refinement) until it is proven to be insufficient. Establishing generative hierarchical causality/multi-scale mechanisms for such simulation models involves 1) identifying a level of abstraction that is "sufficiently complex", 2) utilizing a simulation model characterization strategy that provides the least bias, and therefore greatest explanatory expansiveness, in terms of determining its link to the real-world; this is the Principle of Maximum Entropy [42], and is reflected in calibration methods aimed at finding simulation model parameterizations/configurations that cannot be falsified by existing data.

Thus, the investigatory paradigm in experimental biology generates conditions that preclude the application of traditional means of generating synthetic data to SMMTSD: 1) sparsity of high-dimensional time series data renders any assumptions regarding the true statistical distribution problematic, 2) the generative hierarchical causal relationships inherent to SMMTSD are precluded from the use of formal inference to establish those relationships by the Causal Hierarchy Theorem, 3) simulation models representing the SMMTSD cannot be generated from physics-based fundamental laws and 4) the generative rule structures of potential mechanism-based simulations have perpetual epistemic uncertainty regarding biological rules. Because of these factors, methods for generating SMMTSD using either statistical methods or physics-based/ground-truth/*ab initio* simulations cannot be used. Therefore, an alternative approach is necessary, one that accounts for the limits of ANN-based AI systems and the perpetual epistemic uncertainty/incompleteness regarding cellular and molecular mechanisms.

## 4.0 What features must SMMTSD have?

We identify two main classes of issues to address in generating SMMTSD. The first is mitigating the failure modes for ANN AI systems; these establish issues that need to be overcome by a simulation strategy to generate SMMTSD:

1. ANNs fail to generalize. This is due to ANN training data sets not being effectively and sufficiently representative enough of possible data configurations in the real world. In these cases, the discrepancy between the training set and the eventual application in the real world leads to the condition termed data drift [43-45]. Parenthetically, overcoming this is exactly the goal of using synthetic data for ML/AI, but with the recognition that such synthetic data must be generated in a fashion that provides a more comprehensive, expansive representation of the real-world data (which is the point of this paper). Therefore, the means of generating SMMTSD should be as expansive as possible, representing as much of the breadth of possible configurations of the target/real-world system, so the generative function learned by the ANN is applicable to the widest possible circumstances that might exist in the real world (thereby mitigating and forestalling data drift).

2. Inability to discriminate. A significant number of AI tasks involve distinguishing between one group from another. As such, there is an initial supposition that there is an actual difference present between groups via a distinguishable phenotype/outcome. The issue with much existing time series data is that the range of values within each group is nearly always larger that the difference between some statistically determined characteristic of each group (be it mean or median). Since the statistical distributions are not known, the only way to create synthetic data that represents each different group is to generate it mechanistically, given the criteria of non-falsifiability by existing data and expansiveness in terms of potential explanations given a particular hypothesis structure (see following below regarding "Epistemic Uncertainty").

Overcoming these two issues intrinsic to ANNs are therefore key goals for the construction and use of a putative mechanism-based multiscale simulation approach to generating SMMTSD. Combining these ANN-limit mitigation requirements with the previously defined need to deal with perpetual epistemic uncertainty in simulation model structure leads to the following proposed requirements, with accompanying rationale, below:

1. Choosing a "sufficiently complex" abstraction level for the multiscale mechanism-based simulation model. All computational models incorporate parameters within the rules/equations that make up the model. In mechanism-based multi-scale simulation models of biological processes those rules often represent cellular functions and molecular events, such as receptor binding, signaling, gene activation, protein synthesis or secretion, etc. However, these models do not explicitly represent every component of every step present in the cell; in practice this is impossible because the sum-total of interactions between components, or even the total set of components, is not known. However, it is possible to construct such models where the unknown features (equivalent to latent variables) can be structured such that these features represent (primarily) the *responsiveness* of the represented functions in the model. We therefore consider a "sufficiently" complex multi-scale mechanism-based model as one where all essential cellular behavioral functions for a given purpose of the models are represented via a selection of chosen pathways, but with a latent space of variables that represent the differential responsiveness/gain of the represented functions (representing in aggregate unknown/unidentified genetic, epigenetic or signaling pathways not explicitly represented in the model). If such a model structure is followed, then the unknown features/latent variables can be aggregated into a multi-dimensional set of configurations that are responsible for generating the data heterogeneity seen across biological populations [46]. Therefore, the next step is developing a means of operating on this model in a way that maximizes the expansiveness of behavioral representation used to generate SMMTSD.

2. Maximizing the expansiveness of the generated SMMSTD to minimize data drift for the ANN. It is readily evident that there can be no preconceived means of restricting the possible configurations of these latent responsiveness features. What is needed is a means of minimizing potential a priori bias in the representation of this latent space. The concept of minimizing bias by maximally quantifying ignorance is the fundamental basis for the Principle of Maximum Entropy, a formal method that is grounded in Information Theory and Statistical Physics [42]. While our proposed approach does not directly and formally apply the Principle of Maximum Entropy to synthetic data generation, the general principle of not restricting the range of possible system configurations by some means outside the available data is crucial to mitigate overfitting during training of the ANN. A key difference in our proposed approach is that we operate from the starting point of a knowledge-based hypothesis structure that is embodied by the simulation model and utilize the latent space of unrepresented features, represented in a mathematical object, as the mean of maximizing ignorance/entropy (in an information theoretic sense).

3. Translating expansiveness into an alternative view of calibration. In general practice, the process of calibrating a computational model involves finding the minimally sufficient set of parameterizations able to "best" replicate an existing data set. Because the space of parameterizations is functionally infinite, there is an intrinsically reductive nature to this task; this reductive paradigm is also manifest in common practice of fitting to a mean of what is in reality a highly variable data set [4-7]. If the goal of the modeling exercise is to provide some insight into the differences between different cohorts manifesting different phenotypes of interest (e.g. disease versus not disease), then this aggregating process is useful and beneficial. However, if the goal is to then use that simulation model to produce synthetic data that captures the total amount of expressiveness in that data set (e.g. the heterogeneity of different trajectories manifesting as the variability seen in the data), then the aggregating/mean-fitting approach is insufficient. Rather, the "noisiness" of the data is exactly what must be encompassed by any calibration process used to describe the latent space of unrepresented features in a fashion that maximizes information-entropy. Additionally, the outlier values in a biological data set cannot be considered the min-max values for that particular variable/feature/molecule. Given the perpetual sparsity of these data sets, there can be no supposition that one may have happened to stumble across the actual maximal value biologically possible for that condition. Rather, the "outliers" in a data set need to be considered as only points within a larger potential distribution able to be generated by a proposed simulation model. Thus, any means of generating SMMTSD must be able to expand the range of possible values for any feature in a biologically plausible fashion.

Therefore, the generation of appropriate SMMTSD for training an AI system requires a multi-scale mechanism-based simulation model that embodies an initial set of generative components and a means of operating over that knowledge structure that minimizes the bias regarding all the interactions (the latent variable space and interactions, including second, third to *n*-order effects)[42]. The following section presents a demonstration of such an approach, which uses an agent-based model (ABM) constructed in a particular fashion, and a non-traditional conception of model "parameter space" that represents an initial approximation of a means of characterizing the latent space of unrepresented interactions for a given model. Note that this method of using ABMs, which addresses the generative hierarchical nature of SMMTSD, is distinct from those approaches that use ABMs to generate virtual populations for epidemiological studies, where the trajectories of internal states of the agents is not the primary focus [47-49].

## 5.0 A proposed example of generating SMMTSD

Here we present an example of a strategy for generating SMMTSD to train an AI. This example is not intended to imply that the specific steps and methods listed below are the *only* means of meeting the above stated criteria for generating SMMTSD, but does illustrate addressing all the required criteria for such synthetic data. This approach includes:

1. The use of cell-level ABMs. Cell-level ABMs represent biological systems as interactions between different cell types where each cell type is governed by a defined set of literature-based rules (for a recent review of ABMs in biomedicine see [50]). Such ABMs are desirable for synthetic data generation because: 1) they embody knowledge and spatial interactions not readily replicated with a set of differential equations (which would be readily and misleadingly be reconstituted by training an ANN), 2) they incorporate stochasticity at a generative level (where it exists in biology) and therefore are able to produce the non-Gaussian stochastic distributions seen at the system level and 3) are able to encompass epistemically uncertain or undefined biological features if constructed in a particular way (see next) .

2. The incorporation of epistemic uncertainty in a mathematical object call the Model Rule Matrix (MRM) (Figure 1) [51]. Any representation of a biological mechanism consists of two components: the aspects of that biological mechanism (e.g. a pathway) that is explicitly represented, and a set of effects that are acknowledged to exist (given the inter-connectedness of biological systems) but are not known or explicitly represented. We assert that the vast majority of biological heterogeneity arises from modifications of the strengths/responsiveness/reactivity of a finite set of biological rules. The MRM accounts for these factors by representing the sets of potential interactions that constitute the unknown, latent set of effects.

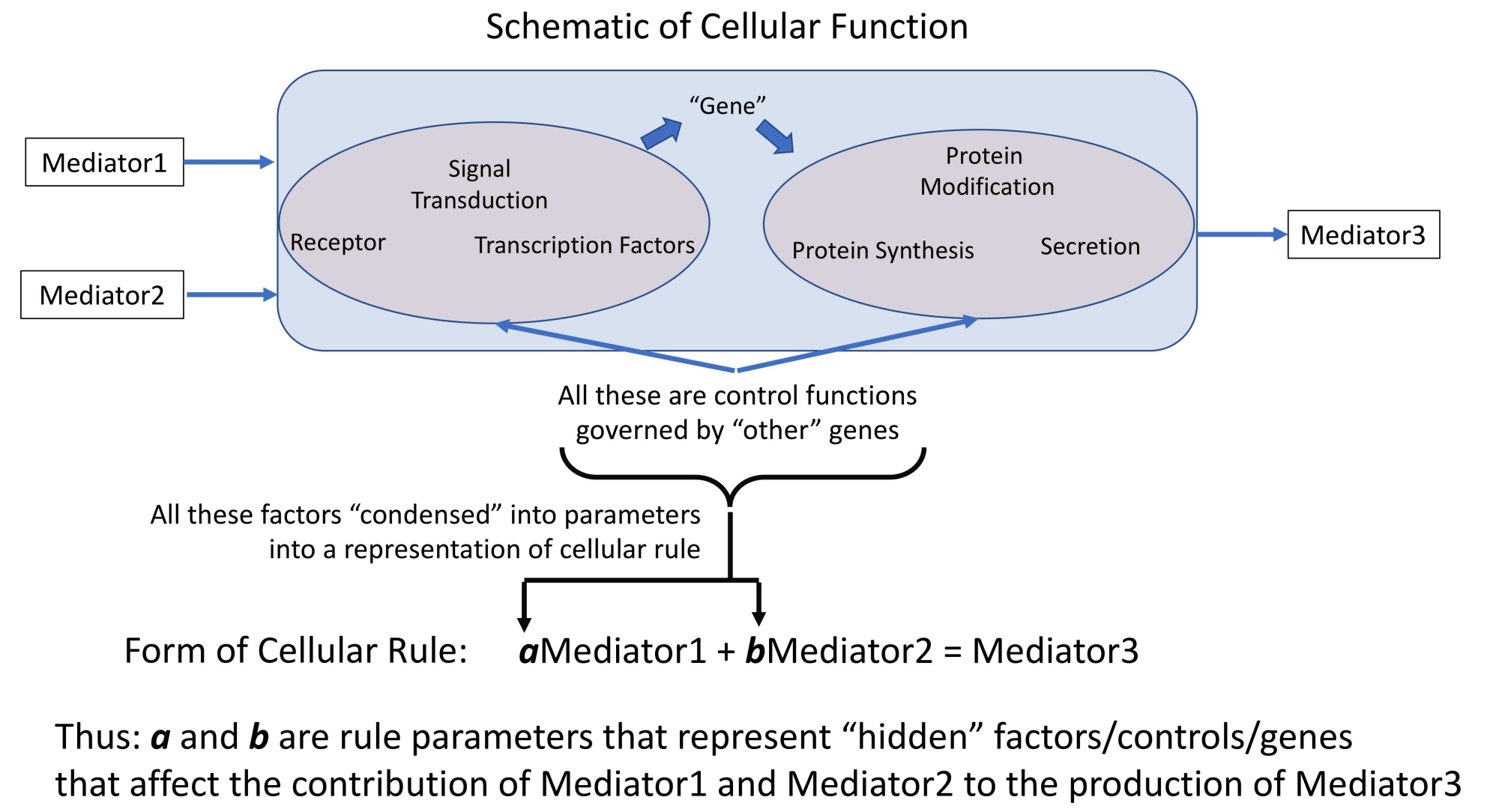

**Figure 1:** Depiction of how representation of cellular behavioral rules governing the effect and generation of various mediators is accomplished by rule parameters. Cellular rules are presented as input-output relationships for specific cell types; in practice not every mechanistic step is represented in such a rule. The weight of each contributing mediator to the overall function of the cell is represented by the parameters associated within each rule. We pose that these parameters essentially aggregates the influence of non-represented or unknown cellular components on the represented rule. Reprinted from Ref 51 under the Creative Commons License.

3. Incorporating the concept of non-falsifiability in the generation of an unbiased, expansive synthetic data set. We have developed a ML pipeline that uses AL to intelligently search across the MRM parameter space to identify the boundaries of that space that encompass a time series data set [52]; this space thereby represents the maximal set of non-falsifiable interactions (consistent with the information theoretic concept of entropy [42]) within the system. This meets the expansiveness criteria needed to overcome the limits of generalizability in training an AI ANN.

The sum effect of this pipeline is the generation of a SMMTSD that is grounded on a known, putative knowledge structure, yet incorporates the sum-total of unknown interactions in that knowledge structure, thereby obscuring the underlying generative model from the AI ANN as it trains on it.

We present an example output of this approach using a well-established ABM of systemic inflammation, the Innate Immune Response ABM (IIRABM) [53, 54], to create a synthetic cytokine trajectories expanding data from a cohort of trauma patients to distinguish between those that develop acute respiratory distress syndrome (ARDS) from those that do not [55]. The IIRABM was calibrated to a clinical data set from The Uniform Services University/Walter Reed National Medical Military Center of 199 trauma patients, 92 of which developed ARDS at some point during the course of their hospitalization, matched with 107 controls that did not develop ARDS. Data elements that were used to calibrate the model include two primary elements: 1) vital signs/laboratory observables necessary to determine SOFA score; for the respiratory compartment, this consists of the partial pressure of oxygen, complete information regarding respiratory support, and blood oxygen saturation, representing aggregate organ function; and 2) time-series blood-serum cytokine profiles consisting of Interleukin-1□ (IL-1□), Interleukin-1 receptor antagonist (IL-1ra), Interleukin-6 (IL-6), Interleukin-4 (IL-4), Interleukin-8 (IL-8), Interleukin-10 (IL-10), Granulocyte Colony Stimulating Factor (GCSF), Interferon-□ (IFN□), and Tumor Necrosis Factor-□ (TNF□), sampled periodically for the duration of the patient's hospitalization. Simulations of the IIRABM were used to identify parameterizations (which represent varied individuals responding to varied insults) that could not be falsified by the available data, and therefore represented the most expansive potential interpretation of the data given the structure of the IIRABM and its corresponding MRM. Using the nested GA parameter discovery and AL parameterization boundary identification method described in [51, 52, 55] we generated SMMTSDs for those patients that would develop ARDS and those that would not. Representative SMMTSD trajectory spaces can be seen for TNF$\alpha$ (as a representative cytokine) in Figure 2 (republished from [55]. Note several features present in these plots that we have noted as important characteristics for SMMTSDs to be used to train ANNs.

1. Note that the raw data (seen as discrete blue or red points) are highly variable, highly overlap, and have variable and shifting distributions over the course of the time series. This pattern is ubiquitous in time series mediator data, and precludes the ability to make any informed decision on the statistical distribution of such data (which would be necessary for either a statistical generation of synthetic data, or to inform a non-traditional noise function in a stochastic differential equation).
2. While there is considerable overlap in the trajectory spaces for non-ARDS (blue) and ARDS (red), there are distinct points at which these spaces separate. But these points of separation are not necessary present at a real-world sampling point, and only become evident in the synthetic data.

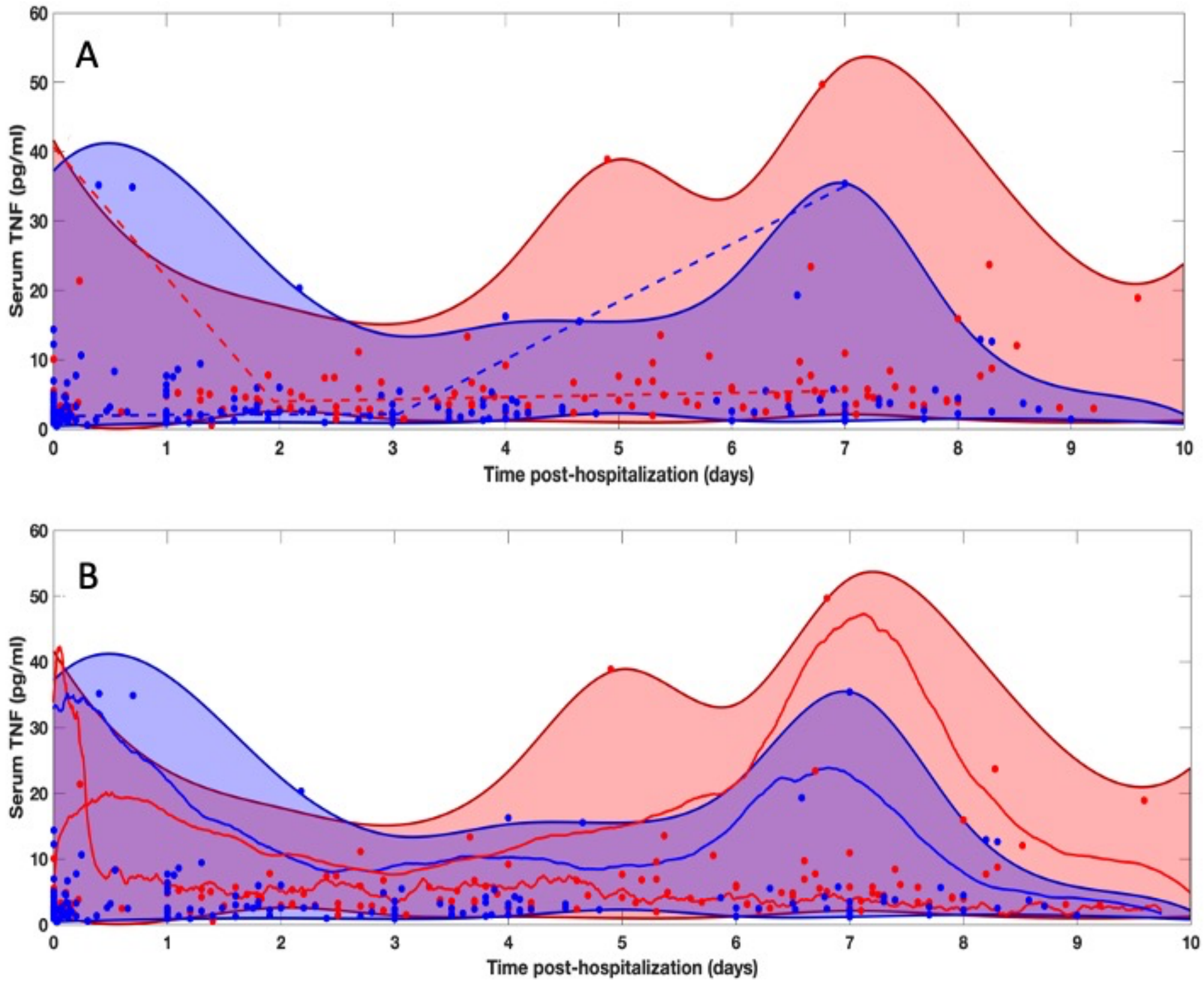

**Figure 2: Comparison of Experimental and Simulated Protein Concentration Trajectories.** In Panel A, we show both simulated and clinically collected blood-serum cytokine measurements for TNFα. Distinct points are the clinically collected data, with red points representing patients that developed ARDS sometime between days 7 and blue points representing patients that did not. The dotted lines indicate representative clinical trajectories seen in the data; note specifically the sparsity of time points for an individual. The shading indicates the boundaries of the model trajectory space for the parameterizations that generate ARDS (red) or not (blue). Note significant overlap between the two spaces given the overlap of data points. However, the key point is that differential parameterizations are able to identify clear regions that are unique to each group. In Panel B, the three solid lines (two red lines eventually developing ARDS and the blue line not) show actual simulated trajectories of TNFα blood-serum concentrations; note here the continuous nature of the trajectories, which would more accurately reflect the actual underlying biological behavior. Reprinted from Ref 55 under the Creative Commons License.

3. The solid blue and red lines in Panel B show examples of specific simulated trajectories that make up the trajectory space (also note the difference between these projected trajectories and the clinical sequential time series represented by the dotted lines in Panel A). Note that these trajectories are affected both by their parameterized MRM and the stochastic processes included in the IIRABM, but it is exactly these trajectories that are being sought by the AI ANN (if the purpose of the AI requires the ANN to know how the system works). This is a key point that separates this approach to generating synthetic time series data from synthetic/virtual populations, where aggregate outcomes are the target output, as is the case for epidemiological investigations (45-47), or approaches that create distinct virtual populations by fitting the mean mediator values (4-7).

## 6.0 The importance of useful failure: iterative refinement

Given the known effects of data drift [43-45] and underspecification [56] on the degradation of AI/ML system performance in real-world applications, for mission-critical applications (such as many potential biomedical applications), it is important to mitigate the consequences of performance degradation by anticipating, as much as possible when performance falls below some identified threshold. As such, the ability to determine the regimes of applicability of a given model is essential. A primary rationale for the use of synthetic data is preempt the performance degradation by increasing the training set to increase the robustness of the trained AI system (mitigating data drift), but this presupposes that the means of generating the synthetic data does not merely accentuate any bias present in the data used to generate the synthetic data. If a data-centric approach is used to generate synthetic data to train ANNs, the identification of bias or insufficient representation in generating the training synthetic data will not become evident until the ANN fails in its intended use. This is because there is no step that allows for a "reality check" of the synthetic data for molecular time series because no one knows what it is "supposed" to look like, for the reasons of perpetual data sparsity and lack of means of predefining the shape of the data distribution. This is distinct from use-cases where such a reality check can be performed, e.g. image analysis or natural language processing, where the synthetically generated data can be examined for believability. Conversely, SMMTSD generated by cell-behavior mechanistic multi-scale simulation models can provide an intermediate step in the applicability of the generated SMMTSD can be evaluated using the criteria of non-falsifiability to determine if data drift of the synthetic data set is occurring. The ability to perform simulation experiments allow the mechanistic models to be refined through both the process of evolution of a MRM-like parameter space characterization, or modification of the underlying simulation model rules based on new biological knowledge. This iterative refinement process allows for "useful failure," since the simulation models are transparent with respect to their composition and can be interrogated to determine where their insufficiencies may be.

## 7.0 Conclusion

In this paper we pose that there is a specific class of data, namely multidimensional molecular-mediator time series data, that presents challenges to traditional means of generating synthetic data to be used for training ANN/AI systems. The specific properties of this data that preclude traditional statistical/GAN methods are:

1. Perpetually sparse data, such that the Curse of Dimensionality cannot be overcome.
2. Such data is explicitly generatively hierarchical, which then imposes the limits of the Causal Hierarchy Theorem on the ability of data centric approaches to identify such structures.
3. The variability/"noise" present in this data does not follow an established distribution pattern (e.g. non-Gaussian because it does not meet the criteria for application of the Central Limit Theorem) and is in fact variable over the course of the time series itself.

We further assert that while mechanism-based simulations are needed to overcome the above restrictions, there are specific properties that are to be met to overcome the limits of traditional simulation means of generating synthetic data:

1. The mechanism-based simulation models need to incorporate a means of dealing with the perpetual epistemic uncertainty regarding knowledge of cellular-molecular mechanisms (e.g. no *ab initio* physics-based modeling).

2. The means of utilizing these simulation models to generate SMMTSD must maximize the generalizability of the generated synthetic data set, and this means maximizing the expansiveness of the applied parameter space, motivated by the Principle of Maximal Entropy.
3. As part of maximizing the expansiveness of the SMMTSD, this means that the simulation method must represent the space of individual trajectories that cannot be falsified by the data. This is because the ultimate goal of the developed/trained ANN/AI is to forecast individual trajectories and/or identify specific controls for individual trajectories.

On one hand, the type of synthetic data covered in this paper is very specific, but on the other hand, generating and using this type of data is a central and overwhelming goal of biomedical research. The generation and use of increasingly granular molecular information about how biological systems work is the primary paradigm in experimental biology, and, in an applied sense, the entire drug development pipeline. Given the importance of such data and the reasonable interest in applying cutting edge ML and AI methods to these problems, it is crucial to develop a means to generate bioplausible SMMTSD to train these systems. The key descriptor here is "bioplausible," as we recognize that that method will not necessarily be a representation of the *fundamental truth*. As noted above, biomedical research operates in a space of epistemic uncertainty and incompleteness, with every hypothesis subjected to the possibility (and perhaps inevitability) of falsification. Rather, the operational strategy in being able to use biomedical knowledge involves being able to find levels of representation that can be demonstrated to be useful. In this sense, the relationship between our proposed methods and reality is analogous to the case of Newtonian Mechanics. The classical understanding of mechanics is broadly applicable to everyday life and has significant predictive power when in its applicable regime (i.e., not quantum or relativistic); however, Newtonian Mechanics alone cannot explain much of the observed motion of our universe. Similarly, while biomedical models suitable to generate SMMTSD have a broad range of explanatory potential, there will certainly be regimes in which the model that generates that data is invalid as these models are, and will be for the foreseeable future, informed by experimental data and not first principles (hence the importance of iterative refinement and "useful failure.").

Thus, it can be concluded from our argument that while the acquisition of more data is certainly necessary, it is not sufficient to meet tasks that require knowledge of the mechanistic processes that cross causal hierarchies (e.g. cell/molecule to individual). It also provides guidelines as to what mechanism-based simulations must account for to be suitable for ANN training. We have presented a specific example that utilizes a cell-based ABM within a ML-augmented pipeline that maximizes the expansiveness of a SMMTSD, but we make no claim as to the uniqueness of this approach, either in terms of the type of simulation method applied, nor the specific components of our parameter-space characterizing pipeline. However, we hope that by presenting our analysis of the current synthetic data landscape and the specific properties of the type of data the represents a considerable proportion of the output of experimental biology, we can provoke other investigators to work in this area.

**Funding:** This work was also supported in part by the National Institutes of Health Award UO1EB025825. This research is also sponsored in part by the Defense Advanced Research Projects Agency (DARPA) through Cooperative Agreement D20AC00002 awarded by the U.S. Department of the Interior (DOI), Interior Business Center. The content of the information does not necessarily reflect the position or the policy of the Government, and no official endorsement should be inferred.

## References

1. Nikolenko SI. Synthetic data for deep learning. arXiv preprint arXiv:190911512. 2019.
2. Bareinboim E, Correa JD, Ibeling D, Icard T. 1on pearl's hierarchy and. Technical report, Technical Report, 2020.
3. Pearl J. Causal inference. Causality: objectives and assessment. 2010:39-58.
4. Sips FL, Pappalardo F, Russo G, Bursi R. In silico clinical trials for relapsing-remitting multiple sclerosis with MS TreatSim. BMC Medical Informatics and Decision Making. 2022;22(6):1-10.
5. Brown D, Namas RA, Almahmoud K, Zaaqoq A, Sarkar J, Barclay DA, Yin J, Ghuma A, Abboud A, Constantine G. Trauma in silico: Individual-specific mathematical models and virtual clinical populations. Science translational medicine. 2015;7(285):285ra61-ra61.
6. Renardy M, Kirschner DE. A framework for network-based epidemiological modeling of tuberculosis dynamics using synthetic datasets. Bulletin of mathematical biology. 2020;82(6):78.
7. Jenner AL, Aogo RA, Alfonso S, Crowe V, Deng X, Smith AP, Morel PA, Davis CL, Smith AM, Craig M. COVID-19 virtual patient cohort suggests immune mechanisms driving disease outcomes. PLoS pathogens. 2021;17(7):e1009753.
8. Kortylewski A, Schneider A, Gerig T, Egger B, Morel-Forster A, Vetter T. Training deep face recognition systems with synthetic data. arXiv preprint arXiv:180205891. 2018.
9. Puri R, Spring R, Patwary M, Shoeybi M, Catanzaro B. Training question answering models from synthetic data. arXiv preprint arXiv:200209599. 2020.
10. Bhandari N. Procedural synthetic data for self-driving cars using 3D graphics: Massachusetts Institute of Technology; 2018.
11. Silver D, Schrittwieser J, Simonyan K, Antonoglou I, Huang A, Guez A, Hubert T, Baker L, Lai M, Bolton A. Mastering the game of go without human knowledge. nature. 2017;550(7676):354-9.
12. Vinyals O, Babuschkin I, Czarnecki WM, Mathieu M, Dudzik A, Chung J, Choi DH, Powell R, Ewalds T, Georgiev P. Grandmaster level in StarCraft II using multi-agent reinforcement learning. Nature. 2019;575(7782):350-4.
13. Perolat J, De Vylder B, Hennes D, Tarassov E, Strub F, de Boer V, Muller P, Connor JT, Burch N, Anthony T. Mastering the game of Stratego with model-free multiagent reinforcement learning. Science. 2022;378(6623):990-6.
14. Hornik K, Stinchcombe M, White H. Multilayer feedforward networks are universal approximators. Neural networks. 1989;2(5):359-66.
15. Creswell A, White T, Dumoulin V, Arulkumaran K, Sengupta B, Bharath AA. Generative adversarial networks: An overview. IEEE signal processing magazine. 2018;35(1):53-65.
16. Bowles C, Chen L, Guerrero R, Bentley P, Gunn R, Hammers A, Dickie DA, Hernández MV, Wardlaw J, Rueckert D. Gan augmentation: Augmenting training data using generative adversarial networks. arXiv preprint arXiv:181010863. 2018.
17. Chen RJ, Lu MY, Chen TY, Williamson DF, Mahmood F. Synthetic data in machine learning for medicine and healthcare. Nature Biomedical Engineering. 2021;5(6):493-7.

18. Candemir S, Nguyen XV, Folio LR, Prevedello LM. Training strategies for radiology deep learning models in data-limited scenarios. Radiology: Artificial Intelligence. 2021;3(6):e210014.
19. Seah J, Boeken T, Sapoval M, Goh GS. Prime Time for Artificial Intelligence in Interventional Radiology. Cardiovascular and Interventional Radiology. 2022;45(3):283-9.
20. Kelly BS, Judge C, Bollard SM, Clifford SM, Healy GM, Aziz A, Mathur P, Islam S, Yeom KW, Lawlor A. Radiology artificial intelligence: a systematic review and evaluation of methods (RAISE). European radiology. 2022;32(11):7998-8007.
21. Kitamura FC, Pan I, Ferraciolli SF, Yeom KW, Abdala N. Clinical artificial intelligence applications in radiology: neuro. Radiologic Clinics. 2021;59(6):1003-12.
22. McAlpine E, Michelow P, Liebenberg E, Celik T. Is it real or not? Toward artificial intelligence-based realistic synthetic cytology image generation to augment teaching and quality assurance in pathology. Journal of the American Society of Cytopathology. 2022;11(3):123-32.
23. Daniel N, Aknin E, Larey A, Peretz Y, Sela G, Fisher Y, Savir Y. Between Generating Noise and Generating Images: Noise in the Correct Frequency Improves the Quality of Synthetic Histopathology Images for Digital Pathology. arXiv preprint arXiv:230206549. 2023.
24. Galaznik A, Berger M, Lempernesse B, Ransom J, Shilnikova A. PMU8 A SYSTEMATIC APPROACH FOR SYNTHETIC REPLICATION OF CLINICAL TRIAL COHORTS USING RETROSPECTIVE REAL-WORLD AND CLINICAL TRIAL DATA. Value in Health. 2019;22:S250.
25. Zand R, Abedi V, Hontecillas R, Lu P, Noorbakhsh-Sabet N, Verma M, Leber A, Tubau-Juni N, Bassaganya-Riera J. Development of synthetic patient populations and in silico clinical trials. Accelerated Path to Cures. 2018:57-77.
26. Myles PR, Ordish J, Tucker A. The potential synergies between synthetic data and in silico trials in relation to generating representative virtual population cohorts. Progress in Biomedical Engineering. 2023.
27. Hernandez M, Epelde G, Alberdi A, Cilla R, Rankin D. Synthetic data generation for tabular health records: A systematic review. Neurocomputing. 2022.
28. Libbi CA, Trienes J, Trieschnigg D, Seifert C. Generating synthetic training data for supervised de-identification of electronic health records. Future Internet. 2021;13(5):136.
29. Baowaly MK, Lin C-C, Liu C-L, Chen K-T. Synthesizing electronic health records using improved generative adversarial networks. Journal of the American Medical Informatics Association. 2019;26(3):228-41.
30. Venugopal R, Shafqat N, Venugopal I, Tillbury BMJ, Stafford HD, Bourazeri A. Privacy preserving Generative Adversarial Networks to model Electronic Health Records. Neural Networks. 2022;153:339-48.
31. Chin-Cheong K, Sutter T, Vogt JE, editors. Generation of heterogeneous synthetic electronic health records using GANs. workshop on machine learning for health (ML4H) at the 33rd conference on neural information processing systems (NeurIPS 2019); 2019: ETH Zurich, Institute for Machine Learning.
32. Tucker A, Wang Z, Rotalinti Y, Myles P. Generating high-fidelity synthetic patient data for assessing machine learning healthcare software. npj Digital Medicine. 2020;3(1):147. doi: 10.1038/s41746-020-00353-9.
33. Levine S. LIVING HEART: USING PREDICTIVE AI/VR MODELS TO REDUCE UNCERTAINTY IN CARDIOVASCULAR DIAGNOSIS AND TREATMENT. Canadian Journal of Cardiology. 2019;35(10):S79-S80.

34. Peng GC, Alber M, Buganza Tepole A, Cannon WR, De S, Dura-Bernal S, Garikipati K, Karniadakis G, Lytton WW, Perdikaris P. Multiscale modeling meets machine learning: What can we learn? Archives of Computational Methods in Engineering. 2021;28:1017-37.
35. Burton II WS, Myers CA, Rullkoetter PJ. Machine learning for rapid estimation of lower extremity muscle and joint loading during activities of daily living. Journal of Biomechanics. 2021;123:110439.
36. Sharma R, Dasgupta A, Cheng R, Mishra C, Nagaraja VH. Machine Learning for Musculoskeletal Modeling of Upper Extremity. IEEE Sensors Journal. 2022;22(19):18684-97.
37. Verleysen M, François D, editors. The curse of dimensionality in data mining and time series prediction. Computational Intelligence and Bioinspired Systems: 8th International Work-Conference on Artificial Neural Networks, IWANN 2005, Vilanova i la Geltrú, Barcelona, Spain, June 8-10, 2005 Proceedings 8; 2005: Springer.
38. Didi K, Zečević M. On How AI Needs to Change to Advance the Science of Drug Discovery. arXiv preprint arXiv:221212560. 2022.
39. Kolla L, Gruber FK, Khalid O, Hill C, Parikh RB. The case for AI-driven cancer clinical trials–The efficacy arm in silico. Biochimica et Biophysica Acta (BBA)-Reviews on Cancer. 2021;1876(1):188572.
40. An G. Specialty grand challenge: What it will take to cross the valley of death: Translational systems biology,"True" precision medicine, medical digital twins, artificial intelligence and in silico clinical trials. Frontiers in Systems Biology. 2022:5.
41. An G. Closing the scientific loop: bridging correlation and causality in the petaflop age. Science translational medicine. 2010;2(41):41ps34-41ps34.
42. De Martino A, De Martino D. An introduction to the maximum entropy approach and its application to inference problems in biology. Heliyon. 2018;4(4):e00596.
43. Nelson K, Corbin G, Anania M, Kovacs M, Tobias J, Blowers M, editors. Evaluating model drift in machine learning algorithms. 2015 IEEE Symposium on Computational Intelligence for Security and Defense Applications (CISDA); 2015: IEEE.
44. Baier L, Jöhren F, Seebacher S, editors. Challenges in the Deployment and Operation of Machine Learning in Practice. ECIS; 2019.
45. Ackerman S, Farchi E, Raz O, Zalmanovici M, Dube P. Detection of data drift and outliers affecting machine learning model performance over time. arXiv preprint arXiv:201209258. 2020.
46. Cockrell C, An G. Genetic Algorithms for model refinement and rule discovery in a high-dimensional agent-based model of inflammation. bioRxiv. 2019:790394.
47. Popper N, Zechmeister M, Brunmeir D, Rippinger C, Weibrecht N, Urach C, Bicher M, Schneckenreither G, Rauber A. Synthetic reproduction and augmentation of COVID-19 case reporting data by agent-based simulation. medRxiv. 2020:2020.11. 07.20227462.
48. Truszkowska A, Behring B, Hasanyan J, Zino L, Butail S, Caroppo E, Jiang ZP, Rizzo A, Porfiri M. High-Resolution Agent-Based Modeling of COVID-19 Spreading in a Small Town. Advanced theory and simulations. 2021;4(3):2000277.
49. Bissett KR, Cadena J, Khan M, Kuhlman CJ. Agent-based computational epidemiological modeling. Journal of the Indian Institute of Science. 2021:1-25.
50. Sivakumar N, Mura C, Peirce SM. Combining Machine Learning and Agent-Based Modeling to Study Biomedical Systems. arXiv preprint arXiv:220601092. 2022.
51. Cockrell C, An G. Utilizing the heterogeneity of clinical data for model refinement and rule discovery through the application of genetic algorithms to calibrate a high-dimensional agent-based model of systemic inflammation. Frontiers in physiology. 2021;12:662845.

52. Cockrell C, Ozik J, Collier N, An G. Nested active learning for efficient model contextualization and parameterization: pathway to generating simulated populations using multi-scale computational models. Simulation. 2021;97(4):287-96.
53. An G. In silico experiments of existing and hypothetical cytokine-directed clinical trials using agent-based modeling. Critical care medicine. 2004;32(10):2050-60.
54. Cockrell C, An G. Sepsis reconsidered: Identifying novel metrics for behavioral landscape characterization with a high-performance computing implementation of an agent-based model. Journal of theoretical biology. 2017;430:157-68.
55. Cockrell C, Schobel-McHugh S, Lisboa F, Vodovotz Y, An G. Generating synthetic data with a mechanism-based Critical Illness Digital Twin: Demonstration for Post Traumatic Acute Respiratory Distress Syndrome. bioRxiv. 2022:2022.11. 22.517524.
56. D'Amour A, Heller K, Moldovan D, Adlam B, Alipanahi B, Beutel A, Chen C, Deaton J, Eisenstein J, Hoffman MD. Underspecification presents challenges for credibility in modern machine learning. Journal of Machine Learning Research. 2020.